\documentclass{article}

\usepackage{lipsum} 
\usepackage{amsmath} 
\usepackage{subcaption}

\usepackage[utf8]{inputenc} 
\usepackage[T1]{fontenc}    

\usepackage{float}

\usepackage{hyperref}       
\usepackage{url}            
\usepackage{booktabs}       
\usepackage{amsfonts}       
\usepackage{nicefrac}       
\usepackage{microtype}      
\usepackage{xcolor}    

\usepackage{cleveref}

\usepackage{graphicx}
\usepackage{xr}
\usepackage{xcite}

\makeatletter
\newcommand*{\addFileDependency}[1]{
  \typeout{(#1)}
  \@addtofilelist{#1}
  \IfFileExists{#1}{}{\typeout{No file #1.}}
}
\makeatother

\usepackage{hyperref}
\usepackage[margin=1in]{geometry}
\crefname{fig}{Fig.}{Figs.}
\Crefname{figure}{Fig.}{Figs.}
\crefname{suppfig}{Supplementary Figure}{Supplementary Figures}
\crefname{extended_fig}{Extended Data Fig.}{Extended Data Figs.}
\usepackage{rotating}

\usepackage{caption}

\DeclareCaptionLabelSeparator{naturebar}{\space|\space}

\usepackage[numbers,sort&compress]{natbib}

\usepackage{xr}
\title{Positioning manuscripts in the scientific landscape with agentic AI\vspace{0.5cm}}
\author{Jiawen Chen$^1$, Zichen Zhang$^1$, Bingxuan Li$^2$, Quan Sun$^{3,4,5}$,\\
Yiyan Zhang$^6$, Edric Tam$^7$, Jinjie Lin$^8$, Didong Li$^{9}$, Yun Li$^{6,9}$, Bingxin Zhao$^{1,10}$\thanks{Correspondence: \texttt{bxzhao@wharton.upenn.edu}.}\vspace{0.2cm}\\
$^1$ Department of Statistics and Data Science, University of Pennsylvania; \\
$^2$ Department of Computer Science, University of Illinois at Urbana-Champaign; \\
$^3$ Center for Computational and Genomic Medicine, Children’s Hospital of Philadelphia; \\
$^4$ Department of Biomedical and Health Informatics, Children’s Hospital of Philadelphia;\\
$^5$ Department of Pathology and Laboratory Medicine, University of Pennsylvania; \\
$^6$ Department of Genetics, University of North Carolina at Chapel Hill; \\
$^7$ Department of Biomedical Data Science, Stanford University; \\
$^8$ School of Management, Yale University; \\
$^9$ Department of Biostatistics, University of North Carolina at Chapel Hill; \\
$^{10}$ Department of Medicine, University of Pennsylvania. \\
}

\begin{document}

\maketitle
\date
\begin{abstract}
Publishing a research manuscript is a routine yet demanding part of scientific life: time-consuming, stressful, and often uncertain in outcome. Recent advances in large language model (LLM)-based agentic AI have shown promise across a range of scientific tasks, and here we ask whether agentic AI can help researchers navigate the publication process itself by reliably inferring a manuscript's eventual publication venue from its content and literature context.
We introduce PASS (Publication-oriented Agentic Scientific System), an agentic system that understands manuscripts within their domain-specific literature context and predicts top-matched publication venues. PASS positions each manuscript within its surrounding literature landscape by reconstructing its local scientific neighborhood, tracing its topic trajectory, and reasoning over field-specific journal spaces.
Evaluated on a leakage-audited benchmark of over 2,000 preprints across 16 biomedical fields, PASS achieved Top-1 accuracy of 50.3\% and Top-5 accuracy of 86.1\%, outperforming state-of-the-art LLM baselines and established journal-selection tools. PASS-produced quality scores, such as impact potential and novelty, aligned with independent measures of publication outcome. We also found that the designed literature retrieval module is the strongest performance contributor, particularly for positioning manuscripts relative to nearby work, and that PASS maintained near-full performance from the abstract alone, whereas LLM baselines required the full manuscript text. An independent human evaluation found strong researcher agreement with PASS's manuscript understanding and recommendation rationale. PASS has been released as a public platform (\url{https://ratemypaper.ai/}) for broad researcher access. 
\end{abstract}

\section{Introduction}
Publishing a scientific manuscript is an essential part of researchers' scientific life. Beyond conducting the study and preparing the paper, researchers often need to decide where the work belongs, how it should be positioned, and whether it meets the expectations of a particular journal. Finding an appropriate journal helps connect the manuscript with editors and reviewers who can best evaluate its merit, limitations, and contribution. These judgments require researchers to interpret journal scope, audience, novelty requirements, editorial priorities, and the related literature~\citep{suiter2019selecting}. They consume substantial time, often involve considerable uncertainty, and can become a source of stress even for experienced authors. A poor match between a manuscript and a journal may delay publication, increase revision time and resubmission burdens, and redirect effort away from actual scientific work~\citep{jiang2019high}. The challenge is particularly pronounced in biomedicine fields, where an increasing volume of manuscripts moves through a broad, heterogeneous, and often slow publication ecosystem~\citep{andersen2021time}.

At the same time, artificial intelligence (AI) is reshaping how scientific work is conducted. Recent large language model (LLM)-based systems can synthesize scientific literature~\citep{asai2026synthesizing,ghareeb2026multi,gottweis2026accelerating}, generate hypotheses~\citep{sanyal2025spark,lu2024ai,ghareeb2026multi,gottweis2026accelerating}, write code~\citep{aygun2026ai}, execute data analyses~\citep{huang2025biomni,lu2024ai,ghareeb2026multi}, draft manuscripts~\citep{lu2024ai,ghareeb2026multi}, and provide review feedback~\citep{lu2024ai,thakkar2026large,zhu2025deepreview}. These advances have encouraged a vision of increasingly automated research workflows, operating as end-to-end engines for scientific research~\citep{kong2026ai,lu2024ai,ghareeb2026multi}. Yet most existing efforts have focused on intermediate research outputs, such as hypothesis generation, study planning, or data analysis. A later-stage AI assistant that researchers need remains largely underexplored: helping researchers navigate the practical question of where a manuscript should ultimately be submitted and published, even as automated workflows increasingly handle much of what comes before it.

This raises the central question we asked here: can agentic AI, which has shown strong performance across a range of scientific and research tasks~\citep{gao2024empowering, kong2026ai}, reliably support researchers in navigating the journal selection process by reasoning from a manuscript's scientific content and its position within the related literature? The task extends beyond simple topic matching. Journal landing depends on multiple interacting factors, including scientific quality, contribution type, evidence strength, scope and audience type~\citep{cagan2013san}, and a manuscript may be suitable for multiple journals that differ in specialization, visibility and expectations for novelty or validation. This distinguishes journal recommendation from much of the recent work on AI-assisted scientific publication, which has mainly focused on computer science conference acceptance and its peer-review process~\citep{zhu2025deepreview,lu2024ai,thakkar2026large} is typically framed as a binary classification decision within a single venue. Journal publication, by contrast, involves selecting among many journals with diverse scopes and standards, whose outcomes are further shaped by editorial decisions, reviewer input, and revision processes, introducing additional uncertainty~\citep{cole1981chance}. As a result, journal recommendation is not a straightforward classification problem; it requires understanding both the manuscript's contribution and how it aligns with different journal environments. Despite this variability, researchers and editors in practice rely on identifiable signals in manuscripts and the surrounding literature to assess journal fit, suggesting that the process, while entailing substantial uncertainty, remains structured enough to model. Building on this observation, existing journal selection tools provide an important foundation for this problem. For example, JANE~\citep{schuemie2008jane} compares the title or abstract of the target paper with publication records and recommends journals that have published similar articles. Pubmender~\citep{feng2019deep} uses word embeddings and a supervised convolutional neural network to classify abstracts into PubMed-indexed journals. These methods are useful but were developed before the emergence of modern LLMs and agentic workflows. They primarily treat journal recommendation as short-text similarity or supervised abstract classification and have not benefited from these recent advances.

Here we introduce PASS (Publication-oriented Agentic Scientific System), an agentic framework for biomedical manuscript understanding, evaluation, and journal recommendation. Its central idea is to situate each manuscript within a domain-specific literature landscape before making a recommendation, drawing on closely related studies to establish immediate scientific context and on reviews and benchmark papers to capture broader field trends, then reasoning over the resulting field-specific journal space. 
The system returns ranked journal recommendations, evidence-grounded rationales, and manuscript-level assessments of novelty, impact, conceptual soundness, and writing quality.
We evaluated PASS on a temporally controlled and leakage-audited benchmark of 2,184 recent bioRxiv and medRxiv preprints linked to their peer-reviewed publication records across 16 biomedical fields. PASS recovered the eventual publication journal with a Top-1 accuracy of 50.3\% and a Top-5 accuracy of 86.1\%, exceeding both state-of-the-art LLM baselines and previously established journal selection tools by a wide margin. In addition, manuscript-level quality assessments produced by PASS were associated with independent measures of publication outcome, indicating that PASS recovers scientific signals extending beyond topical similarity alone. Literature retrieval emerged as the largest contributor to PASS's performance in ablation analyses, particularly for positioning discovery and method papers relative to nearby work. 
Notably, we found that the prediction power is concentrated in the narrative sections, abstract, introduction, and discussion, that define a manuscript's contribution and field position. PASS retained near-full accuracy from the abstract alone, whereas the LLM baselines depended more heavily on full manuscript text. This indicates that a structured, literature-grounded workflow can efficiently extract publication-relevant signals from the sections where researchers most directly frame their question, contribution, and significance.

We further introduce the PASS platform (\url{https://ratemypaper.ai/}), a public-facing system that lets researchers upload manuscripts and receive ranked journal recommendations, evidence-grounded rationales, and supporting literature evidence, intended to assist, not replace, human judgment. An independent human evaluation of the PASS platform by six doctoral-level reviewers across 44 reports found that 88.0\% judged the recommended journal tier close to their expectations and that researchers agreed with 88.7\% of PASS's manuscript-level assessments, confirming that the platform's understanding and rationale are broadly aligned with researchers' expert judgments. In summary, these results establish journal landing as a measurable endpoint of publication-oriented manuscript understanding, demonstrating that agentic AI can help researchers position manuscripts within the scientific landscape.

\section{Results}
\subsection{PASS: an agentic framework for literature-grounded journal recommendation}

\begin{figure}[!h]
    \centering
    \includegraphics[width=\linewidth]{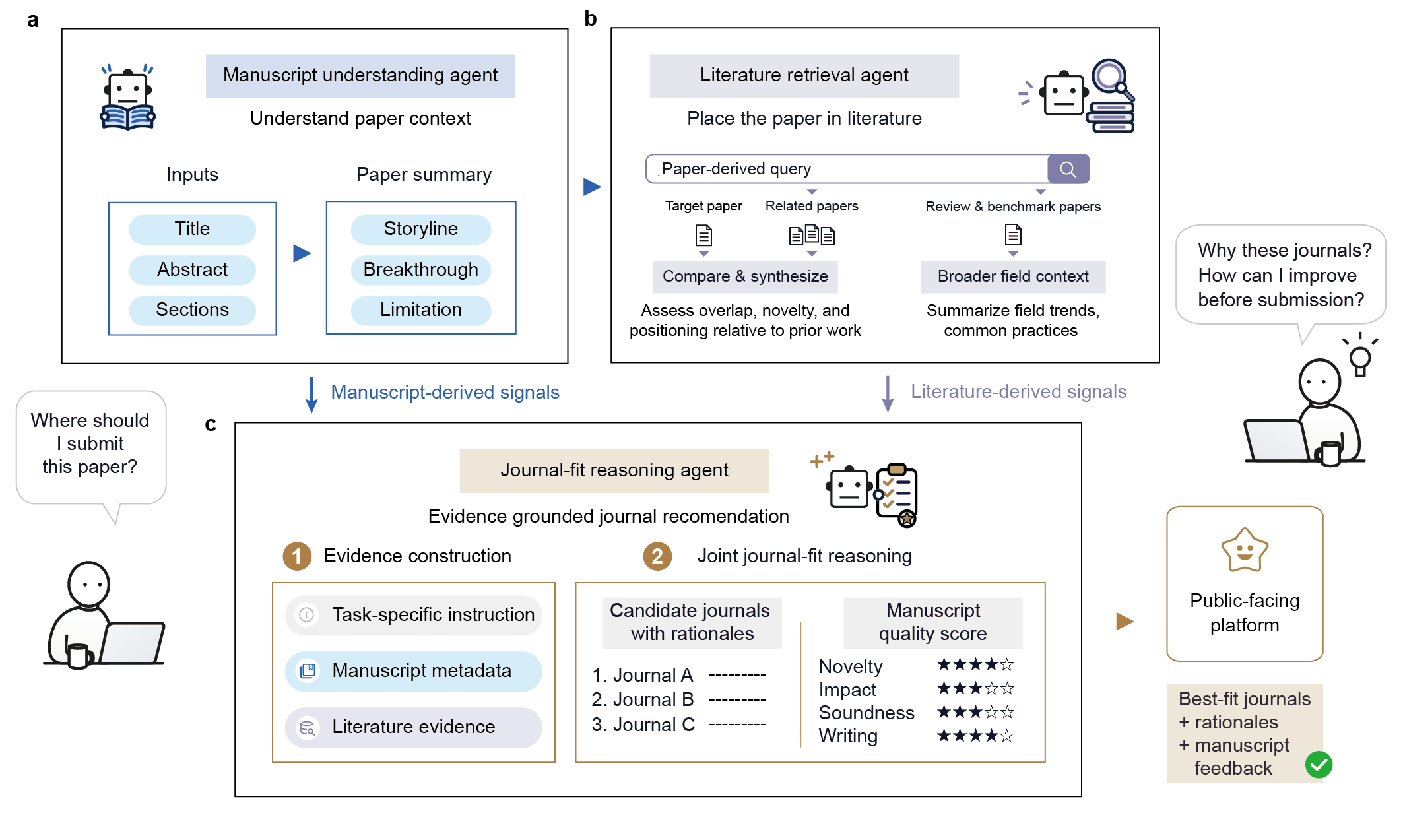}
    \caption{\textbf{Overview of PASS.} \textbf{a}, Taking a paper as input, the manuscript-understanding agent analyzes the title, abstract, and main sections to construct a compact summary of the paper’s storyline, contribution, and limitations, producing manuscript-derived signals. \textbf{b}, The literature-retrieval agent converts the manuscript profile into targeted queries and uses retrieved related papers to assess overlap, novelty and positioning relative to prior work, while review and benchmark papers provide broader field trends. \textbf{c}, The journal-fit reasoning agent integrates task-specific instructions, manuscript metadata and literature evidence to jointly generate a ranked list of candidate journals with rationales and manuscript-level assessments of novelty, impact, soundness and writing quality. The final output provides best-fit journals, evidence-grounded rationales and feedback for strengthening the manuscript before submission.}
    \label{fig:workflow}
\end{figure}

To identify suitable publication venues for a manuscript, we must answer three interconnected questions: What does the paper contribute? How should that contribution be evaluated against the surrounding literature? And which journals are the best fit for its scope, significance, and supporting evidence?
PASS addresses these questions through three coordinated agents: a manuscript-understanding agent that extracts the paper’s core contributions and claims, a literature-retrieval agent that positions them within the relevant scientific landscape, and a journal-fit reasoning agent that evaluates venue compatibility.
Together, these agents transform a submitted manuscript into manuscript-derived signals, literature-derived signals, and evidence-grounded journal recommendations (\Cref{fig:workflow}).

Taking a manuscript as input, the manuscript-understanding agent first converts the submitted paper into a structured representation of its scientific content (\Cref{fig:workflow}a). The agent analyzes the title, abstract, and major manuscript sections to identify the central research question, storyline, contribution type, principal claims, supporting evidence, and key limitations. The resulting representation captures both what the paper studies and how its scientific argument is constructed. It also determines the paper type, distinguishing whether the manuscript is a methodological paper, a discovery paper, or a resource paper, and further differentiates manuscripts that address similar topics but make different forms of contribution, such as a methodological advance, a biological discovery, a benchmark, or a reusable data resource. These manuscript-derived signals provide the foundation for the subsequent assessment and recommendation stages. For journal recommendation, PASS relies primarily on the abstract, introduction and discussion, which we show in Section~\ref{concise} to retain near-full predictive signal relative to the full manuscript. 

The literature-retrieval agent then places the manuscript within its contemporary scientific context (\Cref{fig:workflow}b). Assessments of novelty, importance, and field position cannot be made from the manuscript alone, because they depend on what has already been established and how the work differs from its related studies. Using the structured manuscript representation inferred by the manuscript-understanding agent, the literature-retrieval agent first generates both targeted keyword-based queries and natural language queries that reflect the manuscript's topic, methods, application area, data modality, and claimed contribution. These queries are adapted to the syntax and capabilities of the literature databases used for retrieval, including PubMed, Semantic Scholar and Europe PMC. Specifically, targeted queries incorporate paper-specific keywords (e.g., raw keywords, tissue and disease types, and method names) and boolean operators to improve precision, while natural language queries capture the broader semantic intent of the manuscript to improve recall. By combining these complementary query strategies across different literature databases, the agent ensures comprehensive and contextually relevant literature coverage. The retrieved records are subsequently passed through a two-stage ranking process, where an initial text-embedding-based ranking serves as a pre-filter to identify the most relevant candidates, followed by LLM analysis that assigns each paper a relevance score for more accurate ranking and simultaneously classifies each paper into closely related studies or review and benchmark papers. Closely related papers are further analyzed against the target paper to assess scientific overlap, identify key differences in methodology and findings, and clarify the manuscript's unique contributions and positioning relative to prior work, whereas reviews and benchmark papers provide broader context on field trends, common practices and evaluation standards. Together, these sources form the literature-derived signals used to evaluate the manuscript's unique contribution.

The manuscript-derived and literature-derived signals are assembled into a structured evidence base for the evidence-grounded journal-fit reasoning agent (\Cref{fig:workflow}c). The agent operates within a joint evaluation–recommendation framework that simultaneously assesses manuscript quality and journal compatibility guided by task-specific instructions and the assembled evidence, considering topic alignment, contribution type, scientific breadth, evidence strength, and field position. 
Task-specific instructions define four publication-relevant dimensions: novelty, impact potential, conceptual soundness, and writing quality. These instructions guide the agent to assess manuscripts along each dimension while identifying and ranking journals whose scope, audience, and editorial expectations best match the manuscript.
By coupling manuscript evaluation and journal recommendation within a unified reasoning process, PASS allows each assessment to inform the other: the inferred strength and significance of the work shape the recommended venue tier, while journal-specific expectations provide context for interpreting manuscript quality and fit. Additionally, this approach does not require a separately trained supervised classifier, but instead uses evidence-grounded LLM reasoning to generate coherent, quality-aware journal recommendations together with manuscript-level scores and explanatory rationales.

The final output comprises the five highest-ranked candidate journals by default, evidence-grounded recommendation rationales, and manuscript-level feedback. Rather than presenting journal selection as a black-box ranking, PASS makes each recommendation transparent and inspectable. Researchers can examine why a venue was selected, how the manuscript was positioned relative to the surrounding literature, and which aspects of the work could be strengthened before submission. In this way, PASS supports human-in-the-loop decision-making while leaving the final venue choice to the researcher.

\subsection{PASS improves journal-landing prediction across biomedical fields}

The primary task for PASS is journal-landing prediction. To evaluate this task, we constructed a benchmark using bioRxiv and medRxiv preprints that were later peer-reviewed and published, treating the preprint manuscript as input and the eventual publication journal as the prediction target. To minimize potential knowledge leakage from LLM pretraining, we included only preprints whose first submitted version appeared on or after 1 October 2024, chosen to exclude manuscripts likely to have been present in the model's training data, and used the most recent available version for records with multiple revisions. We used OpenAI GPT-5 as the base model for all evaluated systems~\citep{singh2025openai}.

\begin{figure}[!h]
    \centering
    \includegraphics[width=\linewidth]{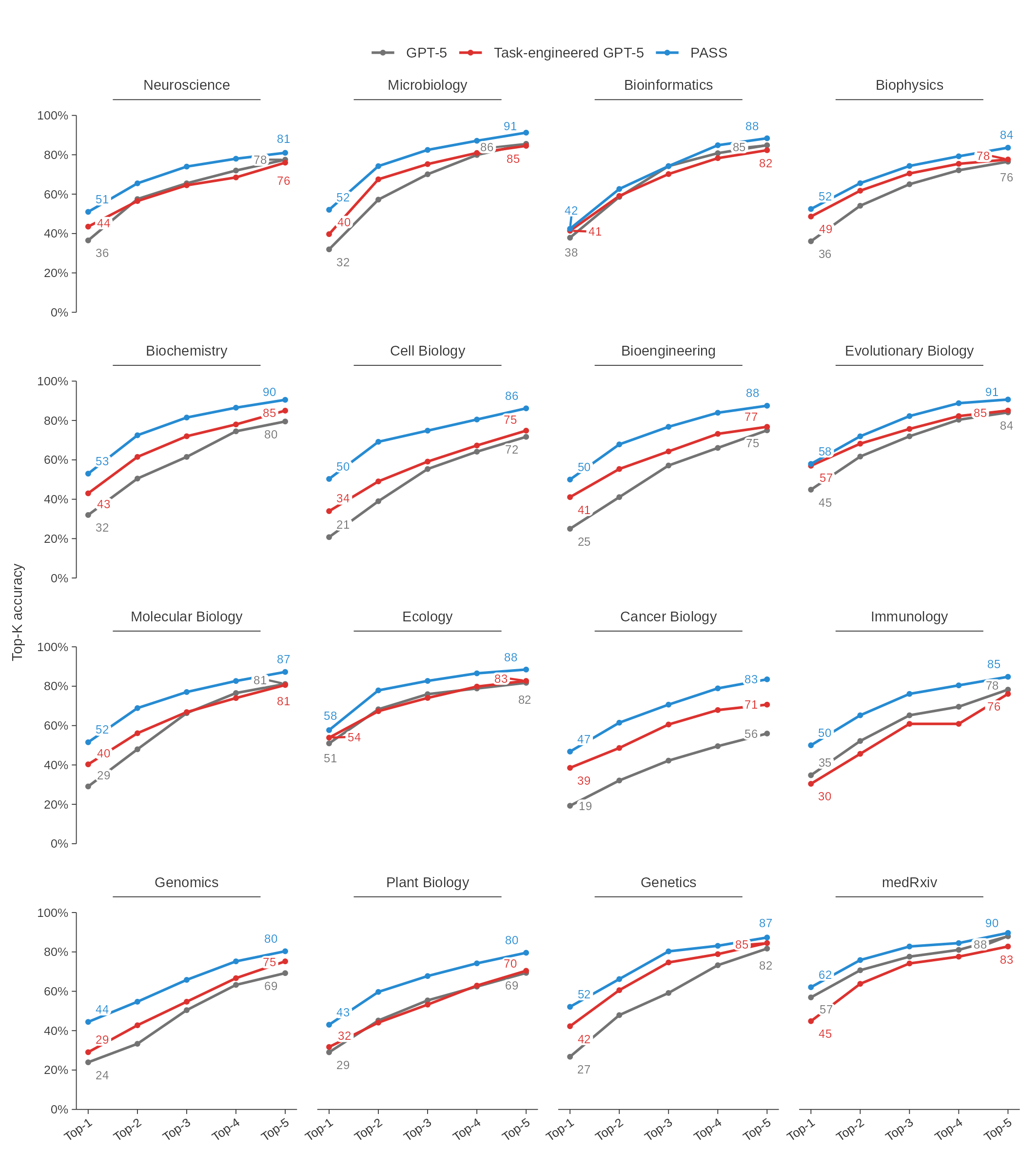}
    \caption{\textbf{PASS improves journal-landing prediction across biomedical fields.} Top-K accuracy of naive GPT-5, task-engineered GPT-5, and PASS across the 15 largest bioRxiv subject categories and a combined medRxiv evaluation set.}
    \label{fig:performance}
\end{figure}

As journal choice is inherently domain-constrained, we evaluated journal landing within field-specific candidate-journal sets. In practice, researchers typically consider a relatively narrow set of venues aligned with their research area and target audience; for example, a biochemistry study would rarely be submitted to a highly specialized immunology journal. Using all biomedical journals as candidates would therefore introduce many implausible candidates that are unlikely to be considered during real-world submission decisions. To construct realistic candidate-journal spaces, we mapped each bioRxiv and medRxiv subject category (\Cref{supp_fig:bio_medrxiv_top15}) to corresponding journal subject areas defined by SCImago~\citep{scimago_sjr}, such as Neuroscience and Medicine. Within each field, we retained papers that were published in leading journals, defined by top SCImago Journal Rank (SJR) within the matched subject area. We evaluated the 15 largest bioRxiv subject categories separately, and a combined medRxiv evaluation panel, as all medRxiv subjects are mapped to the SCImago Medicine category. This procedure yielded 2,184 papers across 16 subject-specific evaluation panels; detailed filtering procedures are provided in the Online Methods. The resulting benchmark evaluates whether, given a manuscript and a realistic field-specific candidate-journal space, a model can identify the journal in which the work ultimately gets published.

We compared three systems on this benchmark. The naive GPT-5 baseline received the manuscript with a simple instruction. The task-engineered GPT-5 received manuscript text together with explicit guidance for reasoning about scope, contribution, novelty, evidence strength, and journal fit. PASS used the full agentic workflow, integrating task-specific instruction, manuscript-derived metadata, and literature evidence. Performance was evaluated by Top-K accuracy (\Cref{fig:performance}), measuring whether the eventual publication journal appeared among the top-ranked recommendations.
Manuscript-only LLMs recovered a substantial journal-landing signal across biomedical fields. The naive GPT-5 achieved an overall Top-1 accuracy of 32.8\% and Top-5 accuracy of 77.6\% across all subjects, far exceeding random selection within large field-specific candidate spaces. This result indicates that biomedical manuscripts contain recoverable information about their likely publication destinations. Prompt engineering further improved the extraction of the signal contained in the manuscript. The task-engineered GPT-5 increased the overall Top-1 accuracy to 41.2\% and Top-5 accuracy to 79.1\%, suggesting that explicit guidance on contribution type, scope, evidence, and audience helps the model better understand journal fit across candidate journals.
PASS achieved the strongest performance across all 16 evaluation panels. The overall Top-1 accuracy increased to 50.3\%, corresponding to gains of 9.1 percentage points over the task-engineered GPT-5 and 17.5 percentage points over the naive GPT-5. Top-5 accuracy increased to 86.1\%, improving by 7.0 and 8.5 percentage points over the two baselines, respectively. PASS achieved the highest Top-K accuracy in every evaluated bioRxiv field and in the combined medRxiv panel. These gains were most pronounced at early ranks, indicating that PASS not only better identifies the most suitable top journals among candidates, but also ranks them more accurately.

We further compared PASS with two established pre-LLM journal recommendation tools, JANE~\citep{schuemie2008jane} and Pubmender~\citep{feng2019deep} (\Cref{supp_fig:jane_pubmender}). PASS outperformed JANE across all 16 field-specific evaluation panels. The comparison with Pubmender was conducted on the Pubmender test split, as Pubmender requires training on abstracts and a predefined candidate space. Despite Pubmender being trained on part of the benchmark dataset, PASS still outperformed it by over 30 percentage points in both Top-1 and Top-5 accuracy (Online Methods and \Cref{supp_fig:jane_pubmender}). This comparison indicates that journal-landing prediction benefits from structured manuscript understanding and literature-grounded reasoning, extending beyond short-text similarity or supervised abstract-based journal classification.
Together, these results establish journal landing as a measurable endpoint of publication-oriented manuscript understanding: biomedical manuscripts contain a recoverable venue signal, task-specific prompting helps extract more of it, and PASS's agentic design recovers still more information beyond what manuscript-only reasoning can access.

\subsection{PASS scores are interpretable and track journal tier}

In addition to accurately predicting journal landing, PASS is designed to make its recommendations interpretable, generating structured, manuscript-level quality scores alongside each journal recommendation. Each report includes an overall score and four dimension-specific scores on a 0-10 scale: impact potential, novelty, conceptual soundness, and writing quality. The overall score summarizes the manuscript's general quality, while the component scores capture different aspects: impact potential reflects expected influence and audience breadth; novelty measures advancement beyond existing literature; conceptual soundness evaluates the coherence of rationale, study design, analytical rigor, and evidentiary support; and writing quality captures clarity, organization, and completeness, including methods and reproducibility details.

\begin{figure}[!p]
    \centering
    \includegraphics[width=\linewidth]{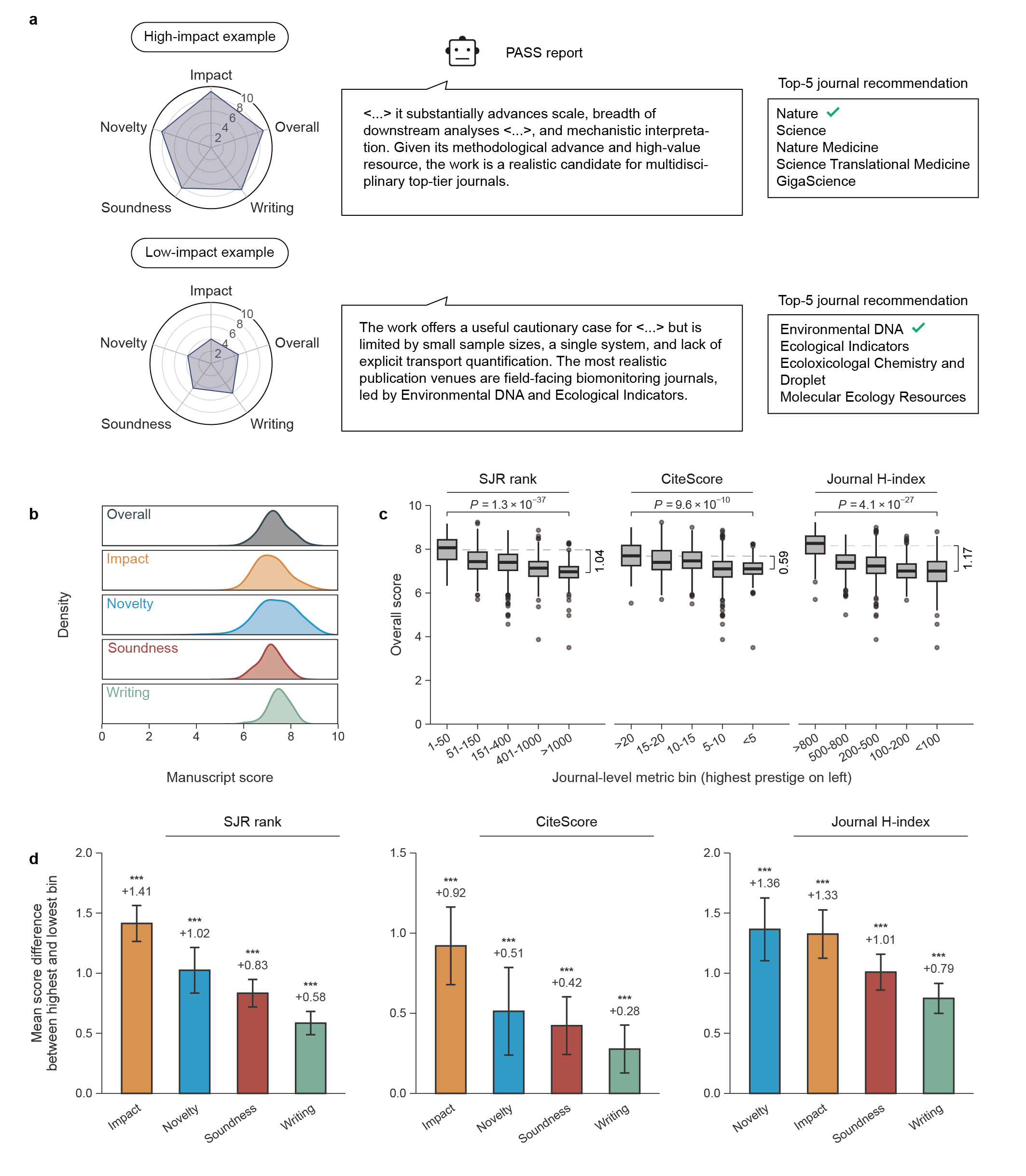}
    \caption{ \textbf{PASS scores provide interpretable manuscript-level signals associated with eventual publication venue.} \textbf{a,} Representative PASS reports for high- and low-scoring manuscripts. The green check indicates the true published journal. \textbf{b,} Distributions of PASS scores across 2,184 evaluated manuscripts. Scores are reported on a 0–10 scale for the overall assessment and four component dimensions. \textbf{c,} Association between the overall manuscript score and three journal-level metrics: SJR rank, CiteScore and journal H-index. Journals are grouped into metric bins, with higher-prestige or higher-visibility bins shown on the left. $P$-values are from a Wilcoxon rank-sum test across the highest and lowest metric bins.  
    \textbf{d,} Dimension-specific score differences between the highest and lowest journal-metric bins. $\star\star\star$ indicates $P$-value $<0.001$ from a Wilcoxon rank-sum test across the highest and lowest metric bins.}
    \label{fig:score}
\end{figure}

For each manuscript, PASS returns this quantitative score profile together with an executive summary explaining how the manuscript's strengths, limitations and positioning within the literature inform the recommended venues. We showcase two representative examples with high and low impact scores, illustrating how these scores guide the recommendation process and link specific manuscript attributes to journal fit (\Cref{fig:score}a). We next examined whether these internal scores capture manuscript-level signals reflected in eventual publication outcomes.

Across the 2,184 evaluated manuscripts, PASS generated well-dispersed score distributions (\Cref{fig:score}b). Median scores ranged from 7.13 to 7.50, with greater variability observed in the impact potential and novelty dimensions. The overall PASS score increased with the visibility and selectivity of the journals in which manuscripts were ultimately published (\Cref{fig:score}c). This trend was consistent across three independent journal-level metrics: SJR rank, CiteScore, and journal H-index (Online Methods). Manuscripts published in the highest-ranked SJR bin received a mean overall score 1.04 points higher than those in the lowest-ranked bin. Comparable patterns were observed for CiteScore and journal H-index, with mean differences of 0.59 and 1.17 points, respectively. These findings indicate that PASS scores capture manuscript-level attributes associated with publication in higher-tier journals, providing quantitative support for the system's recommendations alongside its natural language rationales.

The dimension-specific scores also exhibited a consistent pattern in which higher scores were associated with more selective journals (\Cref{fig:score}d and \Cref{supp_fig:other_score_boxplot}).  
For SJR rank, impact potential exhibited the larger separation between manuscripts in the highest and lowest journal bins, with a mean score difference of 1.41, followed by novelty (1.02), conceptual soundness (0.83), and writing quality (0.58). The same pattern was observed for CiteScore. For journal H-index, novelty (1.36) showed the strongest separation. All four dimensions captured publication-relevant signal across independent journal-level metrics, with the larger separations observed for impact potential and novelty. These results suggest that publication in more selective journals is more closely aligned with expected field impact and originality than with presentation quality alone. Collectively, these findings indicate that PASS recommendations are supported by interpretable, manuscript-level quantitative evidence, with internal scores that align with journal-level metrics.

\subsection{Ablation identifies drivers of journal-landing prediction}

\begin{figure}[!h]
\centering
\includegraphics[width=\linewidth]{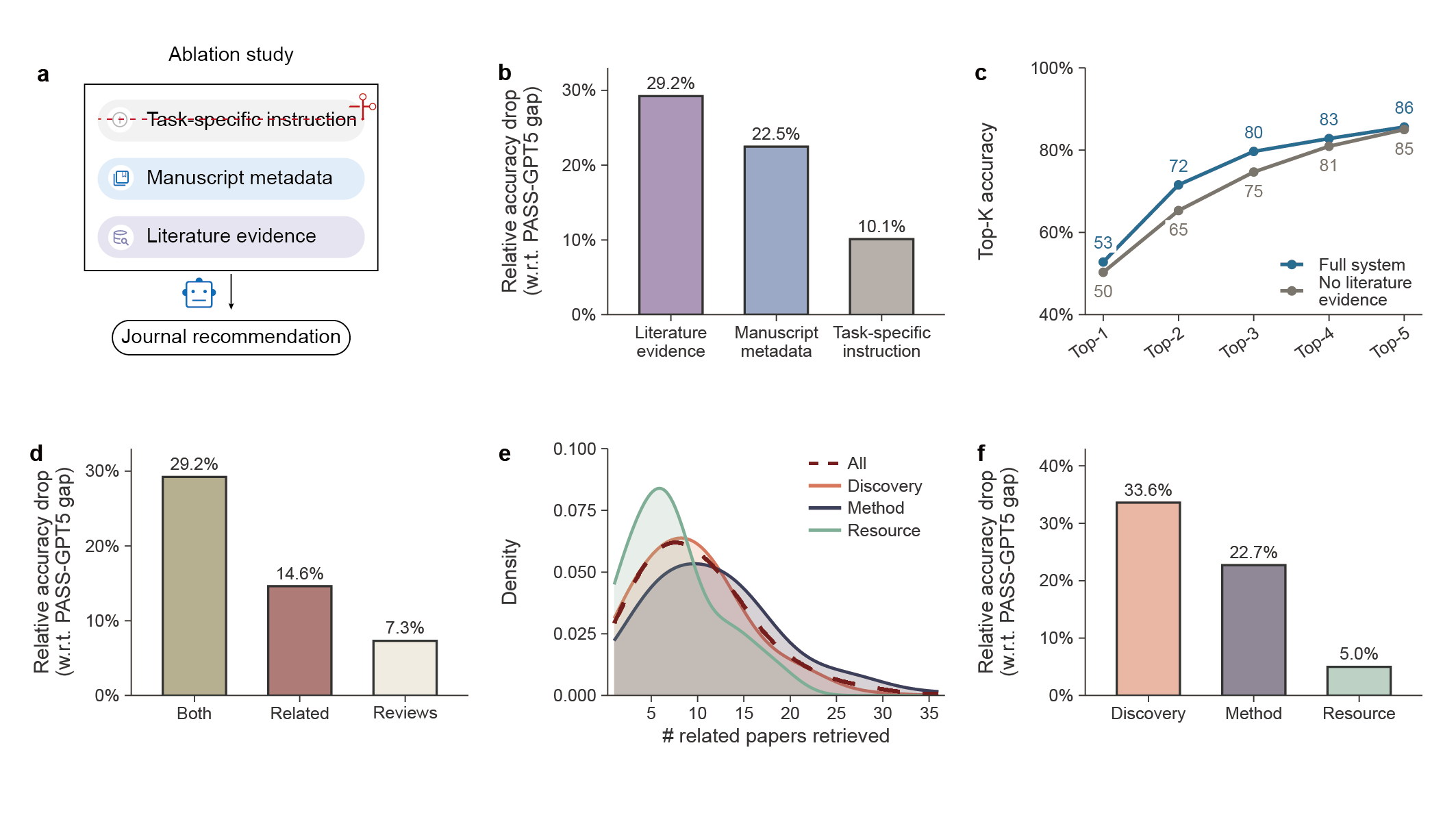}
\caption{\textbf{Ablation analysis identifies complementary signals underlying PASS performance.} \textbf{a,} Ablation design. Ablation variants remove one component at a time. \textbf{b,} Mean Top-K accuracy drop after removing each component. \textbf{c,} Top-K accuracy curve comparing the full system with the ablation variant without literature evidence. \textbf{d,} Retrieval-source ablation showing the effect of removing related studies, review and benchmark papers, or both. \textbf{e,} Distribution of the number of retrieved related papers by manuscript type. \textbf{f,} Mean Top-K accuracy drop after removing literature evidence, stratified by manuscript type.}
\label{fig:ablation}
\end{figure}

After demonstrating that PASS accurately predicts journal landing and provides interpretable, quantitatively grounded recommendations, we systematically dissected the contribution of each system component. PASS frames journal recommendation as an evidence-integration problem solved by a sequence of agent-mediated reasoning steps, combining three complementary sources of information: manuscript-derived signals produced by the manuscript-understanding agent (\Cref{fig:workflow}a), literature-derived signals from the literature-retrieval agent's comparison with prior work (\Cref{fig:workflow}b), and task-specific reasoning instructions that guide journal-fit assessment (\Cref{fig:workflow}c). We quantified the contribution of each source through leave-one-component-out ablations, removing one information stream at a time and measuring the resulting change in journal-landing prediction accuracy (\Cref{fig:ablation}a). We summarized performance by mean Top-K accuracy and expressed the effect of each ablation relative to the performance gap between PASS and the task-engineered GPT-5 baseline.

We find that all three components contributed measurable fractions of PASS's performance gain over the task-engineered GPT-5 baseline (\Cref{fig:ablation}b-\Cref{fig:ablation}c)
. Removing literature evidence produced the largest performance drop, accounting for 29.2\% of the PASS--GPT-5 improvement, followed by manuscript signals (22.5\%) and task-specific instruction (10.1\%). These ablations show that PASS's advantage arises from evidence integration across the agentic workflow. Accurate journal prediction depends on combining manuscript-derived signals, external literature evidence, and explicit journal-fit reasoning, each of which contributes distinct information to the final ranking.

As literature retrieval produced the largest ablation effect, we further examined which part of the retrieved papers contributed more to prediction. The literature-retrieval agent organizes retrieved papers into closely related studies, which capture the manuscript’s immediate scientific context, and review or benchmark papers, which capture broader field-level context. Ablating both evidence types eliminated 29.2\% of the PASS--GPT-5 performance gain. Removing closely related studies alone eliminated 14.6\% of this gain, whereas removing review or benchmark papers eliminated 7.3\% (\Cref{fig:ablation}d). Thus, manuscript-adjacent studies provided the stronger signal, but the larger loss observed when both evidence types were removed indicates that local comparisons and broader field-level context contribute complementary information for positioning the target paper within the literature.

Literature evidence was not equally informative across manuscript types. During manuscript understanding, PASS assigns each manuscript to a broad contribution category, including discovery, method, and resource papers. These categories showed distinct retrieval profiles (\Cref{fig:ablation}e). Across all manuscripts in the ablation analysis, PASS retrieved a mean of 9.93 related papers per manuscript. Method papers had the densest related-literature neighborhoods, with a mean of 11.73 retrieved papers, followed by discovery papers with a mean of 9.35. Resource papers had fewer closely related papers, with a mean of 7.33. 
Consistent with these differences, literature retrieval contributed more strongly to prediction accuracy for discovery and method papers, but had a more modest effect for resource papers (\Cref{fig:ablation}f).
For discovery papers, external evidence helps establish whether a biological finding is new, how it relates to recent work and which audience is most appropriate. For method papers, related studies and benchmarks provide context for assessing technical positioning, methodological novelty, and expected standards of validation. Resource papers, by contrast, appear to depend more heavily on manuscript-intrinsic features, including scale, accessibility, documentation, utility, and potential for reuse.

\subsection{Agentic reasoning preserves journal-landing signal in concise sections}\label{concise}

\begin{figure}[!h]
    \centering
    \includegraphics[width=\linewidth]{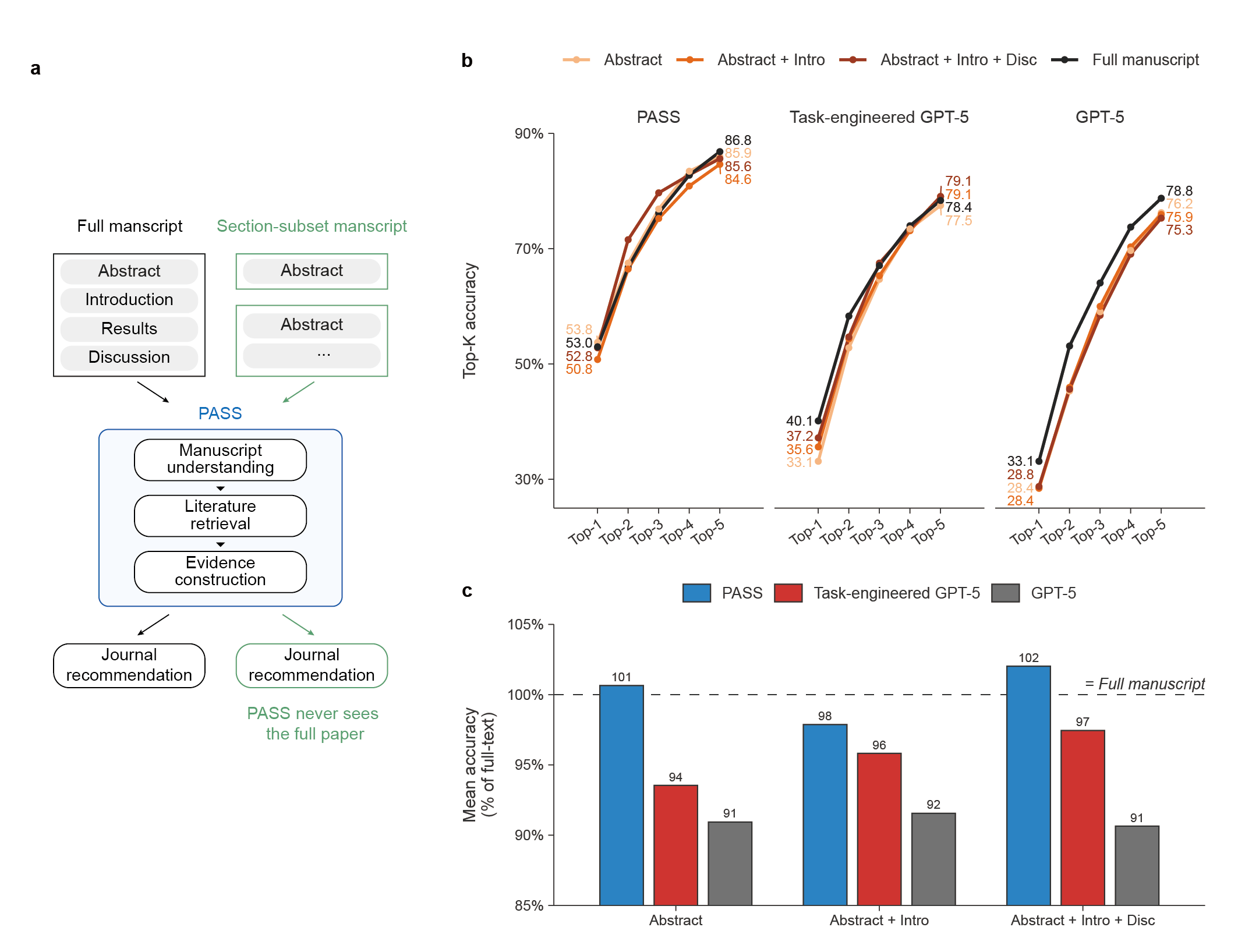}
 \caption{\textbf{Agentic reasoning preserves journal-landing signal from section-subset manuscripts.} \textbf{a,} Evaluation design for PASS using the full manuscript and section-subset manuscripts. \textbf{b,} Top-K accuracy for three models across manuscript input settings. \textbf{c,} Mean Top-K accuracy across (K=1,\ldots,5), normalized to full-text performance.}
    \label{fig:manu_part}
\end{figure}

Initial editorial assessment requires rapid inference of a manuscript's scope, contribution, field positioning, and potential audience. These venue-relevant signals are likely concentrated in certain sections that frame the scientific argument, rather than distributed uniformly across the full manuscript. Typically, the title and abstract provide a concise statement of topic and potential impact, whereas the introduction and discussion clarify motivation, novelty, relationship to prior work, and broader significance. We therefore examined where journal-landing signal is encoded within certain manuscript sections and whether these contribution-framing sections contain enough journal-landing signal for PASS to recover reliable venue predictions without requiring full-manuscript input (\Cref{fig:manu_part}a).

PASS retained near-full performance using only concise contribution-framing sections (\Cref{fig:manu_part}b-\Cref{fig:manu_part}c). With the abstract alone, PASS achieved Top-1 and Top-5 accuracies of 53.8\% and 85.9\%, close to the full-manuscript setting of 53.0\% and 86.8\%. Averaged across K, abstract-only input matched full-manuscript performance (101\% of full-text performance), and adding the introduction and discussion slightly exceeded it, reaching 102\% of the full-text performance. This result suggests that most of the information needed for accurate journal-landing prediction is already contained in the core sections that summarize the study's aims and contributions, allowing PASS to perform effectively without relying on the entire manuscript.

The GPT-5 baselines showed a contrasting dependence on full-text input. For naive GPT-5, all section-subset inputs performed below the full manuscript across Top-K accuracy (\Cref{fig:manu_part}b). Task-engineered GPT-5 was more robust, but its section-subset inputs also remained below the full-text benchmark when averaged across K (\Cref{fig:manu_part}c). Mean Top-K accuracy for task-engineered GPT-5 reached 94\%, 96\% and 97\% of full-text performance using the abstract, abstract plus introduction, and abstract plus introduction and discussion, respectively; for naive GPT-5, the corresponding values were 91\%, 92\% and 91\%. The loss was most pronounced at Top-1: abstract-only input reduced accuracy from 40.1\% to 33.1\% for task-engineered GPT-5 and from 33.1\% to 28.4\% for naive GPT-5. Thus, without literature retrieval and agentic design, GPT-5 baselines required broader full-text context to approach their best journal-landing prediction performance.

Taken together, these results show that the abstract, introduction, and discussion sections contribute most of the journal-landing signal. The agentic design in PASS performs effectively using only these sections, and the GPT-5 baselines likewise draw most of their performance from these same sections, but remain less efficient and need full-text input to close the remaining gap to their best performance. We therefore used the abstract, introduction, and discussion sections as the primary manuscript input for PASS in the main evaluation, while using full-text input for the GPT-5 baselines to provide a conservative comparison. In summary, these findings indicate that journal-landing signal is concentrated in the narrative sections that define a manuscript's contribution and field position, and that agentic reasoning can extract and convert this compact signal into accurate journal recommendations more efficiently than manuscript-only reasoning alone.

\subsection{Human evaluation supports the accuracy and usefulness of PASS reports}

We further assessed the components of PASS that cannot be evaluated against an objective ground truth, including whether the system correctly understands a manuscript, whether its reasoning is scientifically appropriate, and whether its suggested improvements are useful. 
Specifically, we conducted an independent human evaluation using the public-facing PASS platform. Six researchers (all Ph.D. students or Ph.D.s) assessed manuscripts of their choice that were unpublished or only recently published and with which they were either an author or closely familiar. See \Cref{supp_fig:real_report} for a sample report generated by the PASS platform and the survey questions used in the human evaluation.

\begin{figure}[!h]
    \centering
    \includegraphics[width=\linewidth]{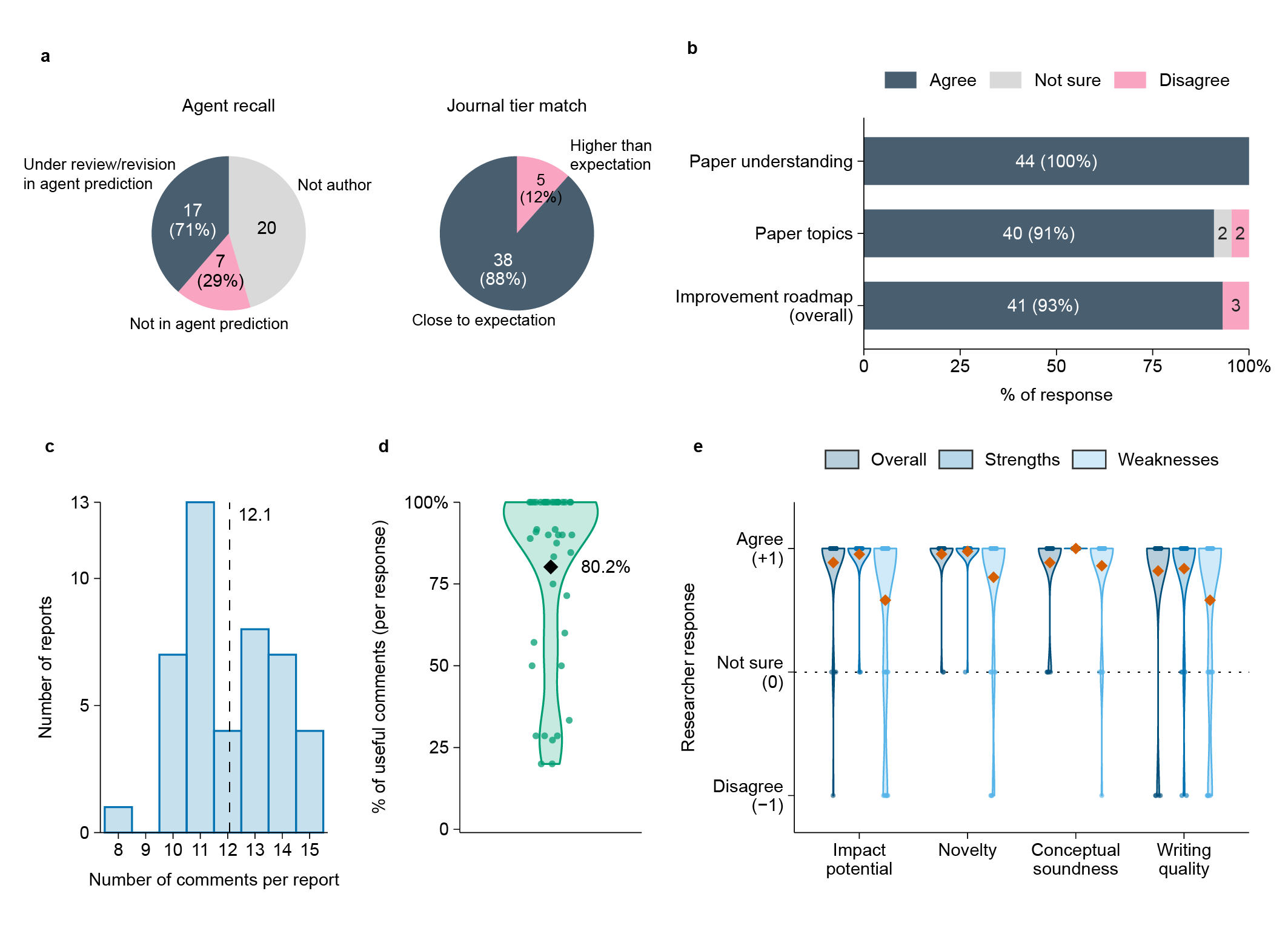}
    \caption{\textbf{Human evaluation supports the accuracy and usefulness of PASS reports.} \textbf{a,} Researcher assessment of PASS journal recommendations. \textbf{b,} Researcher agreement with PASS report quality across paper understanding, topic identification, and improvement roadmap. \textbf{c,} Distribution of the number of manuscript-improvement comments generated per report; the dashed line indicates the mean of 12.1 comments. \textbf{d,} Fraction of comments judged useful by researchers for each report; points indicate individual responses and the diamond indicates the mean of 80.2\%. \textbf{e,} Researcher agreement with PASS manuscript-level assessments across impact potential, novelty, conceptual soundness, and writing quality, evaluated for overall assessment, strengths, and weaknesses.}
    \label{fig:human_survey}
\end{figure}

The researchers evaluated 44 PASS reports spanning 40 unique researcher-uploaded manuscripts. Twenty-two manuscripts were assessed by at least one of their authors, yielding 24 author-verified evaluations for which the intended journal submission trajectory was known. In these cases, the author-reported target journal appeared among the Top-5 PASS recommendations in 17 of 24 reports (71\%) (\Cref{fig:human_survey}a). Researchers were also asked whether the recommended journal tier was higher than, close to, or lower than their own expectation. Among 43 responses, 38 (88\%) judged the recommended tier to be close to expectation. The remaining five rated it as higher than expected; none considered it lower (\Cref{fig:human_survey}a). Thus, PASS generally recommends journals at an appropriate level, and the discrepancies that did occur tended to skew optimistic, not conservative.

Beyond journal recommendations, PASS provided manuscript-improvement roadmaps generated by the journal-fit reasoning agent. Each report provided a structured set of comments targeting the manuscript's positioning, evidence strength, presentation, and journal readiness. We asked researchers to rate the improvement roadmap both at a summary level (overall, \Cref{fig:human_survey}b) and at the level of each actionable item (\Cref{fig:human_survey}c-\Cref{fig:human_survey}d). Across evaluated reports, PASS produced an average of 12.1 comments per manuscript, and researchers judged a mean of 80.2\% of comments per report to be useful. Overall, researchers agreed that the roadmap provided helpful guidance for improving the manuscript in 41 of 44 reports (93\%). These results indicate that PASS not only provides accurate journal recommendations, but also translates its manuscript understanding and journal-fit reasoning into practical feedback that can support strengthening the manuscript before submission.

We additionally evaluated the textual components of each report, including the manuscript-understanding agent's output and the manuscript-level scores and rationales produced by the journal-fit reasoning agent. Researchers agreed that PASS correctly understood the manuscript in all 44 evaluations (100\%), and extracted the correct subject topics in 40 of 44 evaluations (91\%) (\Cref{fig:human_survey}b). 
Researchers further evaluated PASS assessments of impact potential, novelty, conceptual soundness, and writing quality (\Cref{fig:human_survey}e). For each dimension, PASS produced scores and rationales for the overall assessment, strengths, and weaknesses; researchers were asked to rate each as "agree," "not sure", or "disagree." We encoded these responses as $1$, $0$ and $-1$, respectively. Agreement with the overall assessments and identified strengths was consistently high, with mean scores exceeding 0.8 across all four dimensions. Ratings of the identified weaknesses were slightly lower but remained positive, with a minimum mean score of 0.58, indicating that PASS correctly identifies overall assessments and strengths, whereas weakness identification may be more subjective.

\section{Discussion}

Agentic AI is beginning to support many parts of scientific work, including literature review, hypothesis generation, data analysis, experimental planning, and manuscript drafting~\citep{boiko2023autonomous,asai2026synthesizing,ghareeb2026multi,gottweis2026accelerating,sanyal2025spark,huang2025biomni,lu2024ai}. Here we studied whether agentic AI can support an essential practical step of the research process: positioning a manuscript for publication. Publication is not only the endpoint of a research project, it is also a decision about where a scientific contribution belongs, how it should be framed, and which audience is most likely to value it~\citep{suiter2019selecting,gasparyan2013choosing}. PASS addresses this problem by treating journal landing as a measurable endpoint of publication-oriented manuscript understanding: by integrating manuscript understanding, literature retrieval, and journal-fit reasoning, it generated interpretable journal recommendations, manuscript-level score assessments, and revision-oriented feedback, outperforming GPT-5 baselines and established journal-recommendation tools~\citep{schuemie2008jane,feng2019deep}. These results suggest that journal-landing prediction is not simply a problem of topical matching, but a form of evidence-grounded reasoning over a manuscript and its scientific context.

A central insight from this study is that literature context is essential for reliable journal-landing prediction. The eventual journal of a manuscript depends not only on what the paper reports, but also on how its contribution compares with the literature, for example, related studies, field standards, and established journal audiences. This finding is consistent with the broader motivation for retrieval-augmented systems: external evidence can improve factual grounding, provenance, and adaptability when reasoning over knowledge-intensive tasks~\citep{lewis2020retrieval,izacard2021leveraging,gao2023retrieval}. In PASS, literature retrieval produced the largest ablation effect, and closely related papers and review or benchmark papers contributed complementary information. Effective journal-fit reasoning therefore requires reconstructing the manuscript's scientific neighborhood, in which nearby studies help define novelty and competitive positioning, whereas broader reviews and benchmarks capture field-level expectations and the journal landscape.

We also find that much of the journal-landing signal is concentrated in the sections that frame the scientific contribution. The abstract, introduction, and discussion encode the research question, motivation, novelty, relationship to previous work, and broader significance. PASS recovered near-full performance from these concise sections, whereas GPT-5 baselines required full-text input to achieve their best performance. Thus, the agentic design improves not only accuracy but also information efficiency. Furthermore, because PASS depends on only these concise sections rather than a complete manuscript, it can support researchers much earlier in the research process. 
Specifically, by using the parts of the manuscript most relevant to journal fit and amplifying them through retrieval and structured reasoning, PASS enables publication-oriented feedback well before a manuscript is complete, even from abstract-only input.
At an early stage of research, researchers could draft a short summary to clarify the question, expected contribution, and intended audience; PASS could use this early framing to provide literature-grounded feedback on how the proposed work is positioned, what prior studies it should consider, and which aspects of the evidence would be valued by the audience. In this way, the system can support scientific development from the earliest framing of a question onward, not only the final journal-selection decision. By using PASS, authors can explore alternative framings of a research question, identify missing context, and refine the study's contribution before experiments are conducted and the full manuscript is written.

To translate publication-oriented agentic reasoning into a practical research tool, we implemented PASS as a public-facing interactive platform. For each uploaded manuscript, the platform generates a report that explains how the manuscript was understood, what literature evidence was retrieved, why particular journals were recommended, and how the manuscript could be improved before submission. The human evaluation served as an initial beta test of the platform, where researchers were able to upload manuscripts of their choice, inspect the reports, assess the quality of manuscript understanding, and judge whether the improvement roadmap was useful. These results support the intended role of PASS as decision support for researchers, not as an automated substitute for editorial judgment. Its value lies in making journal-fit reasoning transparent, inspectable, and actionable, consistent with broader calls for accountable and human-centered use of AI in scientific workflows~\citep{gao2024empowering,wang2021human}.

To our knowledge, this is the first work to establish journal landing as a measurable, benchmarked task for agentic AI systems. Looking ahead, several directions could extend this framework. While PASS achieved strong performance, each agent in the workflow stands to benefit from more advanced design as AI techniques continue to advance rapidly; we intend to continue developing the public-facing platform to incorporate these advances as they emerge.
Another limitation of the current benchmark is that we only observe the final publication journal. For each manuscript, we do not know the full submission trajectory.  For example, which journals the paper was submitted to; if rejected, whether it was desk-rejected or rejected after review; and how it was revised before publication. In practice, publication is a sequential decision process under time pressure: each unsuccessful submission can delay publication and reduce the perceived novelty of the work. By releasing PASS as a public-facing platform, future work can collect richer prospective data on submission trajectories, revision comments, and publication outcomes. Such data could support more adaptive systems that learn from real submission trajectories and provide increasingly personalized, context-aware guidance. More broadly, PASS points toward a more proactive role for agentic AI in scientific publication. Beyond recommending journals, future systems could help researchers monitor domain trends, identify emerging topics, detect missing comparisons, benchmark against field standards, and refine the framing of a study while the work is still developing. The goal is not to optimize manuscripts for prestige, but to help researchers deliver their contributions clearly, select appropriate audiences, and reduce avoidable mismatch between manuscripts and journals. By linking manuscript understanding to journal positioning, PASS points to a broader role for agentic AI in science: supporting not only how research is written, but also how it is communicated, evaluated, and returned to the scientific record as the starting point for future discovery.

\section*{Acknowledgments} 
This research was supported in part by Analytics at Wharton, Wharton AI \& Analytics Initiative, Wharton Dean’s Research Fund, Wharton Postdoc Matching Fund, Perelman School of Medicine CCEB Innovation Center Grant, and the University Research Foundation Grant. This research was also supported by the National Institute of Mental Health under Award Number R01MH136055, and National Institute on Aging under Award Numbers RF1AG082938 and R01AG085581. The content is solely the responsibility of the authors and does not necessarily represent the official views of the National Institutes of Health. We would like to thank the research computing groups at the Wharton School and the University of Pennsylvania for providing computational resources and support that have contributed to these research results.


\section{Online Methods}

\subsection{PASS agentic system details}

\textbf{Overview.} PASS was implemented as a sequential agentic system for journal-landing prediction. The workflow consists of three coordinated agents: a manuscript-understanding agent, a literature-retrieval agent and a journal-fit reasoning agent. The system takes as input a single manuscript file, provided as either JATS XML or PDF, together with optional metadata such as submission date and a candidate-journal table. It returns a ranked list of five recommended journals, probability estimates, journal-specific rationales, manuscript-level scores, literature-positioning evidence and a prioritized manuscript-improvement roadmap. In this study, all language-model calls used GPT-5.

\textbf{Manuscript-understanding agent.} For the benchmark evaluation, we used the JATS XML files corresponding to each bioRxiv or medRxiv preprint. A structured parser extracted the title, abstract, ordered section bodies, author list, affiliations, received date, figure count, table count, and reference count. Parsed sections were mapped to standard manuscript components, including introduction, methods, results, discussion and conclusion when present. For the main evaluation, we used the title, abstract, introduction and discussion as the primary manuscript context in PASS.

The manuscript-understanding agent converted this manuscript context into a structured manuscript profile. The profile included the broad contribution type, selected from discovery, method or resource; subject-area categories; a one-sentence storyline; summaries of the methodological contribution, main findings and key limitations; paper-specific keywords; and five to eight venue-agnostic search-query seeds. These query seeds were short phrases designed to capture the manuscript’s topic, methods, application area, data modality and claimed contribution. The resulting manuscript-derived signals were passed to the literature-retrieval agent.

\textbf{Literature-retrieval agent.} The literature-retrieval agent used the query seeds generated by the manuscript-understanding agent to retrieve relevant literature from PubMed, Europe PMC and Semantic Scholar. Queries were designed to capture both closely related primary studies and broader review or benchmark papers relevant to the manuscript topic. We retrieved title, journal, publication time, abstract when available and persistent identifiers such as DOI or PubMed ID. Records retrieved from multiple sources were merged using available identifiers and title-based matching to remove duplicates.

To avoid temporal leakage, retrieved records were linked to the manuscript submission timeline, and only studies available at or before the preprint submission date were used as supporting evidence in the benchmark evaluation. Retrieved papers were then organized into two evidence types: closely related studies and review or benchmark papers. Closely related studies were intended to capture manuscript-adjacent context, whereas review or benchmark papers were intended to capture broader field-level context. Candidate records were prioritized using a two-stage ranking procedure. An initial embedding-based ranking step filtered retrieved records using abstract-text similarity to the manuscript profile. A second LLM-based ranking step performed finer-grained relevance assessment, assigned relevance scores and classified records into the two evidence types. The highest-ranking records from each category were retained as literature-derived evidence for journal-fit reasoning.

In addition, publication records for the first author and last or corresponding author were summarized from PubMed and OpenAlex when available. These summaries included typical publication venues, dominant subject areas, recent publication activity and an h-index proxy. Author context was used only as a secondary tie-breaking signal and was not allowed to override manuscript-quality or journal-fit evidence.

\textbf{Journal-fit reasoning agent.} The journal-fit reasoning agent integrated the structured manuscript profile, date-filtered literature evidence, author-context summary and candidate-journal table. Retrieved evidence items were enumerated so that the model could refer to specific records in its rationales. The prompt instructed the agent to jointly evaluate manuscript quality and journal compatibility, considering topic alignment, contribution type, scientific breadth, evidence strength, novelty, impact potential, conceptual soundness and writing quality. The agent returned a structured JSON object containing the manuscript classification, manuscript-level scores, five ranked journal recommendations, probability estimates, journal-specific rationales, a summary of the manuscript’s literature positioning and a prioritized improvement roadmap. Leakage-control instructions explicitly prohibited reliance on future papers, post-publication records, final publication metadata or direct cues to the eventual journal.

\textbf{Multi-run recommendation aggregation.} To reduce stochastic variation, each assembled prompt was submitted to GPT-5 three independent times using independent sampling. Each run returned a complete structured JSON object and was stored separately for auditing. The three top-five journal lists were then fused using a probability-weighted variant of reciprocal-rank fusion (RRF). For each candidate journal $j$, the aggregation score was

$$
S(j)=\sum_{r=1}^{N}\frac{p_r(j)}{K+\mathrm{rank}_r(j)}, \qquad K=60,
$$

where $N=3$, $\mathrm{rank}_r(j)$ is the one-indexed rank of journal $j$ in run $r$, and $p_r(j)$ is the probability assigned to that journal in the corresponding run. The final PASS output consisted of the five journals with the highest aggregated scores, together with the associated rationales, manuscript-level assessments and improvement roadmap.

Prediction performance was evaluated using Top-(K) accuracy, defined as whether the ground-truth published journal appeared within the first (K) journals in the ranked recommendation list. We report Top-1 to Top-5 accuracy, with Top-5 corresponding to the five journals returned by each system after multi-run aggregation.

\subsection{Naive and task-engineered GPT-5 baselines}

We evaluated two prompt-only GPT-5 baselines using the same evaluation panels, candidate-journal sets and ground-truth publication labels used for PASS. Both baselines called GPT-5 directly, without retrieval, agentic design, candidate-paper evidence or literature-context construction. They differed in prompt structure and in the degree of task-specific guidance provided to the model.

\textbf{Naive GPT-5 baseline.} The naive baseline was designed as a minimal-instruction condition. For each manuscript, the raw JATS XML file was inserted into the user message, followed by the category-specific candidate-journal list and an instruction to predict the top five journals from that list in valid JSON format. The prompt contained no scoring rubric, role assignment, structured manuscript assessment or journal-fit reasoning instructions. Because raw JATS XML may contain preprint identifiers or publication-related metadata, the prompt included a short leakage-control instruction directing the model not to search for or infer publication status from the title, DOI or metadata, and to treat the manuscript as a new unpublished submission. Each manuscript was queried three independent times, and the resulting ranked lists were aggregated using the same probability-weighted RRF fusion procedure used for PASS.

\textbf{Task-engineered GPT-5 baseline.} The task-engineered baseline was designed to approximate a strong single-call GPT-5 journal recommender without retrieval or agentic scaffolding. The model received the raw JATS XML, and the candidate-journal table with SJR-derived journal metrics. The prompt was kept as similar as possible to the journal-fit reasoning prompt used in PASS, so that performance differences primarily reflected the absence of retrieval and multi-step evidence synthesis rather than differences in task wording. The model was instructed to evaluate the manuscript across novelty, impact potential, conceptual soundness and writing quality, with each dimension rated on a 0--10 scale and supported by manuscript evidence before generating journal recommendations. Unlike PASS, this baseline did not receive retrieved candidate-paper evidence, literature-positioning summaries or other retrieval-derived information, and therefore operated only on the manuscript content, author information and candidate-journal list. The same leakage-control instruction used for the naive GPT-5 baseline was included. Sampling settings, three-run evaluation and reciprocal-rank fusion aggregation were identical across the naive baseline, task-engineered baseline and PASS.

\subsection{Benchmark dataset construction}

We constructed the benchmark from bioRxiv and medRxiv preprints whose first version was received between 1 October 2024 and 28 February 2026. This time window was chosen to reduce the risk that manuscript--journal pairs were present in the training data of the base model used in this study. Starting from 119,645 records in the bioRxiv/medRxiv metadata dump, we applied three sequential filters. First, we retained manuscripts with a v1 received date on or after 1 October 2024, yielding 108,631 manuscripts. Second, for manuscripts with multiple versions, we collapsed records by DOI and retained the latest available version, yielding 89,142 manuscripts. Third, we retained only preprints with verified links to subsequent peer-reviewed publications at the dataset freeze date of 19 April 2026, as indicated by non-missing published DOI and published journal fields in the metadata. This yielded 19,734 published preprints. The published journal was used as the ground-truth label for journal-landing prediction.

We stratified the published cohort by subject area using the native bioRxiv and medRxiv category annotations. For each subject area, we constructed a field-specific candidate-journal set using the 2024 SCImago Journal Rank metadata. SCImago assigns journals to one or more subject areas, including Biochemistry, Genetics and Molecular Biology; Agricultural and Biological Sciences; Immunology and Microbiology; Medicine; and Multidisciplinary. To define a realistic journal space for each preprint category, we mapped each category to the SCImago subject area with the greatest overlap among the final publication journals observed in that category. For example, bioRxiv genetics preprints most frequently mapped to journals in Biochemistry, Genetics and Molecular Biology.

For each preprint category, candidate journals were defined as the top 100 journals by SJR within the matched SCImago subject area, combined with journals in the SCImago Multidisciplinary category. This step ensured that broad general-science journals, such as \textit{Nature} and \textit{Science}, were included when appropriate, because they are indexed by SCImago under Multidisciplinary rather than under individual biomedical subject areas. We retained only manuscripts whose eventual publication journal appeared in the corresponding candidate-journal set. For categories with more than 200 eligible manuscripts after this filtering step, we randomly subsampled 200 manuscripts for evaluation. This procedure produced the final benchmark of 2,184 manuscripts across 16 evaluation panels. The benchmark metadata are publicly available on Hugging Face (\url{https://huggingface.co/datasets/jiawennnn/PASS-benchmark}), including preprint identifiers, titles, version information, received dates, published journal labels, published DOI fields, publication dates, candidate-journal sets and journal-level SCImago-derived metrics.

\subsection{Benchmark data leakage audit}

We performed an additional audit to assess whether the journal-prediction benchmark could be affected by direct model recall of manuscript--journal associations. The audit was conducted on the full 2,184-paper evaluation cohort, defined as the papers evaluated by all four prediction pipelines across the 16 subject-area panels. For each manuscript, we provided GPT-5 with the title, the author list and the abstract and the model was asked, under a constrained JSON output schema, whether it could already identify the destination journal from memory, with the field knows restricted to yes or no. Across all 2,184 manuscripts, GPT-5 returned \textit{``knows = no''} for every entry, yielding 0 recalled journal assignments.

\subsection{Comparison with established journal-selection systems}

\textbf{JANE.} We compared PASS with JANE~\citep{schuemie2008jane}. For each manuscript in the benchmark dataset, we submitted the title and abstract to the JANE endpoint. We requested the top 30 suggestions to allow downstream filtering before computing Top-k accuracy.

Two filters were applied to the raw JANE outputs before scoring. First, we removed predictions corresponding to preprint repositories, including bioRxiv, medRxiv, arXiv, chemRxiv, Research Square and SSRN, because JANE indexes some preprint servers as journal-like sources. Second, we removed likely self-matches in which JANE appeared to retrieve the formally published version of the input preprint. For each suggested journal, we inspected the top supporting article returned by JANE and compared its title with the query title. A prediction was flagged as a self-match and excluded if the title Jaccard similarity was at least 0.65, or if the title Jaccard similarity was at least 0.50 together with a JANE-reported similarity score of at least 95\%. After these filters, the remaining ranked list was truncated to the top five journals and stored for evaluation. 

\textbf{Pubmender.} We also compared with Pubmender, a pre-LLM neural journal recommendation model based on abstract-level classification~\citep{feng2019deep}. Since the source code of Pubmender is not accessible, we implemented a PyTorch version of the published architecture. Abstracts were tokenized and truncated or zero-padded to 350 word identifiers, embedded using a learned 200-dimensional embedding layer and passed through three one-dimensional convolutional blocks: Conv1d(200, 256, kernel size 3), ReLU and MaxPool1d(2); Conv1d(256, 128, kernel size 4), ReLU and MaxPool1d(2); and Conv1d(128, 96, kernel size 5), ReLU and MaxPool1d(2). The resulting representation was flattened and projected through fully connected layers with dimensions 512, 128 and $N_{\mathrm{classes}}$, using ReLU activations, dropout of 0.2 after the first two linear layers, cross-entropy loss, the Adam~\citep{kingma2014adam} optimizer and L2 weight decay of $10^{-5}$. Hyperparameters followed the original Pubmender specification where applicable.

Because Pubmender is a supervised neural model, it required task-specific training before evaluation. We trained the model on the published-paper BGM top-100 subset of our benchmark, using 1,819 manuscripts for training, 228 for validation and 228 for testing (8:1:1). Training was performed for 60 epochs on a single NVIDIA V100 GPU using cross-entropy loss and the Adam optimizer, and the checkpoint achieving the highest validation Top-5 accuracy was selected for final evaluation (\Cref{supp_fig:jane_pubmender}b). PASS and JANE were evaluated on the same 228 paper set.

\subsection{Journal-level metrics}

Journal-level metrics were obtained from the 2024 SCImago Journal \& Country Rank release downloaded from \url{https://www.scimagojr.com}. We used a single fixed SCImago snapshot for all candidate-journal construction, ranking and downstream analyses. No journal records were added or updated after the snapshot date, ensuring that candidate-journal tables and reported metrics were reproducible.

For each journal, we retained three fields. The first was SCImago Journal Rank (SJR), a size-independent, prestige-weighted citation measure in which citations are weighted by the influence of the citing journal. SJR was used as the primary metric for ranking journals within candidate-journal tables. The second was \textit{Citations per document, 2 years} (as CiteScore in \Cref{fig:score}), defined by SCImago as the average number of citations received in a given year by citable documents published in the preceding two years. This field was used as a short-window citation-impact metric. The third was journal H-index. A journal has an H-index of $h$ if it has published at least $h$ articles that have each received at least $h$ citations, providing a cumulative measure of citation impact.

\subsection*{Code and data availability}
The PASS platform is freely available at \url{https://ratemypaper.ai/}. The benchmark metadata are publicly available on Hugging Face (\url{https://huggingface.co/datasets/jiawennnn/PASS-benchmark}), including preprint identifiers, titles, version information, received dates, published journal labels, published DOI fields, publication dates, candidate-journal sets and journal-level SCImago-derived metrics.


\end{document}